\documentclass{article} 
\usepackage{iclr2015,times}
\usepackage{hyperref}
\usepackage{url}
\usepackage{graphicx}
\usepackage{caption}
\usepackage{subcaption}
\usepackage{placeins}
\usepackage{multirow}
\usepackage{wrapfig}
\usepackage{footnote}
\usepackage{tablefootnote}
\usepackage{amssymb}

\title{Flattened \\ Convolutional Neural Networks \\ for Feedforward Acceleration}

\author{
Jonghoon Jin, Aysegul Dundar \& Eugenio Culurciello \\
Purdue University, West Lafayette, IN 47907, USA \\
\texttt{\{jhjin, adundar, euge\}@purdue.edu} \\
}

\iclrfinalcopy 

\begin{document}

\maketitle

\begin{abstract}
We present flattened convolutional neural networks that are designed for fast feedforward execution.
The redundancy of the parameters, especially weights of the convolutional filters in convolutional neural networks has been extensively studied and different heuristics have been proposed to construct a low rank basis of the filters after training.
In this work, we train flattened networks that consist of consecutive sequence of one-dimensional filters across all directions in 3D space to obtain comparable performance as conventional convolutional networks.
We tested flattened model on different datasets and found that the flattened layer can effectively substitute for the 3D filters without loss of accuracy.
The flattened convolution pipelines provide around two times speed-up during feedforward pass compared to the baseline model due to the significant reduction of learning parameters.
Furthermore, the proposed method does not require efforts in manual tuning or post processing once the model is trained.
\end{abstract}

\section{Introduction}
Recent success on fast implementation of convolutional neural networks (CNNs), and new techniques such as dropout enable researchers to train large networks that were not possible before. 
These large CNNs show great promise in visual and audio understanding which make them useful for applications in autonomous robots, security systems, mobile phones, automobiles and wearable supports. 
These applications require networks with high degree of accuracies, but also networks that can be executed in real-time.
However, CNNs are computationally very expensive and require high performance servers or graphics processing units (GPUs).

To accelerate forward and backward passes of CNNs, there has been extensive work for efficient implementation of CNNs on GPUs \citep{krizhevsky2012imagenet, chetlur2014cudnn} and CPUs \citep{vanhoucke2011improving}, including linear quantization of network weights and inputs.
For mobile platforms, like smartphones, computation of these big networks is still demanding and takes place on off-site servers because of their limited computing power and battery life.
However, that requires a necessity to a reliable connectivity between the mobile device and off-site servers.
Because this is not the case always, custom architectures have been explored for power and speed efficient implementation of CNNs \citep{jin2014efficient, merolla2014neuroIBM}.

Another approach to speed up evaluation of CNNs is to reduce the number of parameters in the network representation.
The work by \citet{denil2013predicting} is a good example to show that these networks have high redundancy in them.
Considering that state of the art CNNs require hundreds of filters each layer and consist of three to five convolutional layers in general, finding essential representation with smaller parameters brings significant performance boost in terms of time and memory.
\citet{jaderberg2014speeding, denton2014exploiting} exploit the redundancy within convolutional layer after training and could obtain speedup by keeping the accuracy within $1\%$ of the original models.

In this work, we take a similar approach to decrease the redundancy of the filters in convolutional neural networks.
We achieve that in the training phase by separating the conventional 3D convolution filters into three consecutive 1D filters: convolution across channels (lateral), vertical and horizontal direction.
We report similar or better accuracies on well-known datasets with the baseline network which has about ten times more parameters.

\section{Related Work}
Convolutional Neural Networks (CNNs) exhibit high redundancy in the representation expressed as weight and bias parameters.
The filters, visual interpretation of weights, in the network often have similar patterns and some of them have noise rather than distinct features.
Having redundancy in the parameters not only degrades learning capacity of networks but accompanies unnecessary computations during feedforward pass as well as backpropagation.
Many approaches have been proposed to find compact representation of CNNs by applying constraints on cost function or structure.

Sparsity in filters often help accelerate computations of filtering operation by simply skipping computations over non-zero values.
Sparse feature learning aligned with findings in V1 neurons is proposed by \citet{lee_2007_sparse_code}.
By iterating $L_1$ and $L_2$ regularizer, this work successfully finds sparse and essential basis filters in over-complete system.
However, size of the non-zero values in sparse filters is irregular and these non-zero values are located in arbitrary positions.
Due to the arbitrary locations and shapes of sparse filters, in practice it is difficult to take advantage of sparsity with highly parallelized processing pipelines.

A classic but powerful method to accelerate filtering operations is to condition separability on object function \citep{rigamonti_2013_separable_cvpr} so as to force the network to learn separable filters.
In the literature, difficulties of optimization problem with $L_1$ penalty are relaxed.
The separable 2D filter has a rank of one and makes the operation equivalent to two consecutive 1D convolutions, which significantly shortens the evaluation time of CNNs by an order of magnitude.

Recent works from \citet{jaderberg2014speeding, denton2014exploiting} speed up CNNs evaluation time with low rank filter approximation.
They compress convolutional layer of pre-trained networks by finding an appropriate low-rank approximation.
\citet{denton2014exploiting} extends the method to a large-scale task.
Pre-trained 3D filters are approximated to low rank filters and the error is minimized by using clustering and post training to tune the accuracy.
Their method demonstrates speedup of convolutional layers by a factor of two, while keeping the accuracy within $1\%$ of the original model.

Another approach to reduce the parameters in CNNs is to explore the connections between layers.
In the structure of state of the art CNNs, all convolutional planes are fully connected.
Such connection scheme can handle and generalize all possible cases in training but in practice it is hard to learn sparse connectivity in output prediction.
The importance of sparse connections in CNNs has been mentioned in the recent work \citep{szegedy_2014_inception}.
Also the connectivity previously was investigated by \citet{culurciello_2013_connection}, though many issues remain open for further research.

We apply structural constraints to conventional CNNs in order to learn 1D separated filters for feedforward acceleration.
Our method does not alter training procedure of CNNs; backpropagating the error from output to the input along constrained paths.
The approach bypasses difficulties in optimization problem witnessed in \citet{rigamonti_2013_separable_cvpr}, but successfully learns 1D convolution filters.
The proposed method does not require any manual tuning or changes in the structure once trained \citep{jaderberg2014speeding, denton2014exploiting}, which simplify overall method.
The concept of 1D convolution across channels is equivalent to the operation denoted as \textit{mlpconv} layers in Network in Network \citep{lin_2013_nin}.
We also use $1\times1$ filters across channels to increase model discriminability for local patches within the receptive field.
This layer also determines the number of filters in the subsequent layers because we add vertical and horizontal 1D filters which only performs convolution per channel.
The difference between their and our work is that all of our filters are one-dimensional which provides significant reduction in parameters.

\section{Flattening convolution filters}

\begin{figure}
  \centering
  \begin{subfigure}[b]{0.3\textwidth}
    \includegraphics[width=\textwidth]{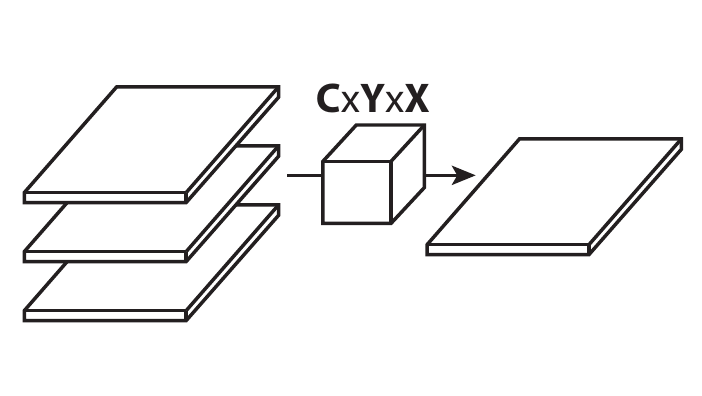}
    \caption{3D convolution}
    \label{fig:illust-conv-3d}
  \end{subfigure}%
  \qquad
  \begin{subfigure}[b]{0.6\textwidth}
    \includegraphics[width=\textwidth]{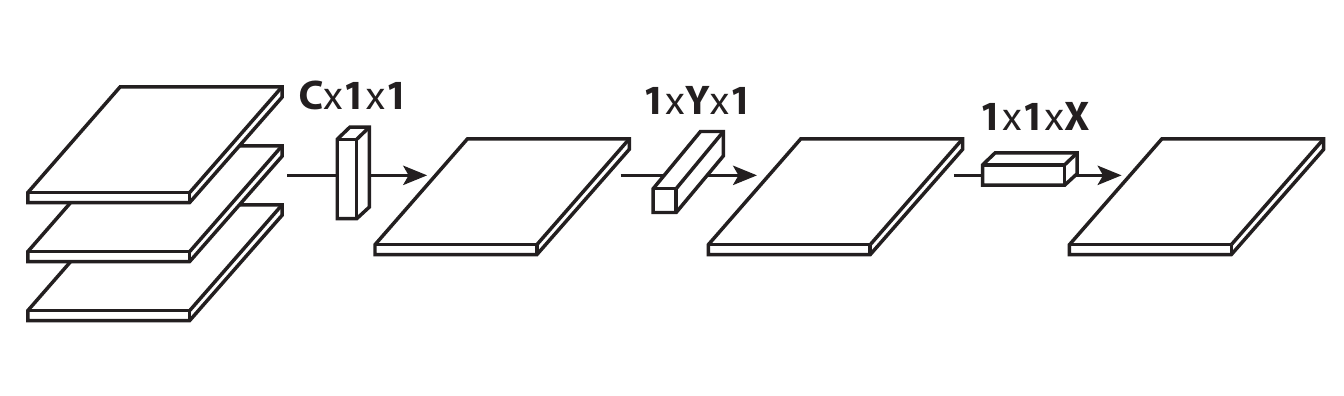}
    \caption{1D convolutions over different directions}
    \label{fig:illust-conv-1d}
  \end{subfigure}
  \caption{The concept of 3D filter separation under rank-one assumption in the context of CNNs.
           Summation over all planes convolved with 2D filters produces a single output plane, which can be considered as 3D convolution.
           Three consecutive 1D filtering is an equivalent representation of 3D filter if its rank is one.
           $C$ is the number of planes and its value is $3$ in the diagram.
           $Y$ and $X$ denote filter height and width respectively.
           Bias is not considered here for simplicity.}
    \label{fig:comparison-filters}
\end{figure}

Similar to the notation used in \citet{denton2014exploiting}, weights in CNNs can be described as 4-dimensional filters:
$W \in \mathbb{R} ^{C \times X\times Y\times F}$, $C$ is the number of input channels, $X$ and $Y$ are the spatial dimensions of the filter, and $F$ is the number of filters or the number of output channels.
Convolution for each channel output requires a filter $W \in \mathbb{R} ^{C \times X\times Y}$ and is described as
\begin{equation}
  F_f(x,y) = I \ast W_f= \sum_{c=1}^C \sum_{x'=1}^X \sum_{y'=1}^Y I(c, x-x', y-y') W_f(c, x', y')
  \label{eq:conv-vanilla}
\end{equation}
assuming a stride of one where $f$ is an index of output channel, $I \in \mathbb{R} ^{C \times N\times M\times F}$ is the input map, $N$ and $M$ are the spatial dimensions of the input.

A rule of thumb to accelerate multi-dimensional convolution is to apply filter separation.
Under rank-one assumption of the filter $W_f$, the unit rank filter $\hat{W_f}$ can be separated into cross-products of three one-dimensional filters as follows.
\begin{equation}
  \hat{W_f} = \alpha_f \times \beta_f \times \gamma_f
  \label{eq:separability}
\end{equation}
We denote 1D convolution vectors as a lateral filter $\alpha_f$ : convolving features across channels; vertical filter $\beta_f$ : across $Y$ dimension; horizontal filter $\gamma_f$ : across $X$ dimension.

However, separability of filters is a strong condition and the intrinsic rank of filter $W_f$ is higher than one in practice.
As the difficulty of classification problem increases, the more number of leading components is required to solve the problem \citep{montavon_dcnn_kernel_2011}.
Learned filters in deep networks have distributed eigenvalues and applying the separation directly to the filters results in significant information loss.

Alternatively, we could restrict connections in receptive fields so that the model can learn 1D separated filters upon training.
When applied to the equation \ref{eq:conv-vanilla}, a single layer of convolutional neural networks is modified to

\begin{equation}
  \hat{F_f}(x,y) = I \ast \hat{W_f} = \sum_{x'=1}^X  \left( \sum_{y'=1}^Y  \left( \sum_{c=1}^C  I(c, x-x', y-y') \alpha_f(c)\right) \beta_f(x')\right)  \gamma_f(y')
  \label{eq:conv-flattened}
\end{equation}
With this modification number of parameters to calculate each feature map decreases from $XYC$ to $X+Y+C$, and number of the operations needed decreases from $MNCXY$ to $MN(C+X+Y)$.

Here we define flattened convolutional networks as CNNs whose one or more convolutional layer is converted to a sequence of 1D convolutions.
We did not add the bias terms in the equations above to keep the equations clean.
However; bias term is important for the training, and removing the bias terms in some of the 1D filters results in very slow learning.
In our tests, we have separate bias terms for each three 1D filters.

\section{Experimental Results}

We tested the performance of the proposed model in alignment with a baseline model of CNNs on different classification tasks.
In experiments, we used the Torch7 environment \citep{collobert_torch7_2011} to demonstrate model performance as well as to handle customized gradient updates.

\subsection{Training Baseline Model}
\label{sec:train-baseline}
We choose a CNN model architecture same as the baseline model used in \citet{srivastava2013discriminative} with a smaller multilayer perceptron.
We keep the structure of CNNs to be generic so as to minimize unwanted interruption from hidden variables and make comparison to flattened model transparent.
The model  consists of 3 convolutional layers with $5 \times 5$ filters and double stage multilayer perceptron.
The number convolutional filters in each layer  are $96$, $128$ and $256$ respectively, each layer includes a rectifier linear unit and a max-pooling with sizes of 2 and strides of 3. 
The two fully-connected multilayer perceptrons have $256$ units each.
The model is regularized with five dropout layers with the probability of $0.1$, $0.25$, $0.25$, $0.5$ and $0.5$ respectively from lower to higher layer in order to prevent co-adaptation of features.

In our tests, we did not use data augmentation in order to concentrate learning capacity of models with respect to its structure.
The training is initially set up with a learning rate of $0.1$ and a momentum of $0.9$.
After the first eight epochs, the learning rate is reduced by one tenth.
With the vanilla CNN and training configuration we were able to achieve performance comparable to state-of-the-art results on many datasets (see Table \ref{table:acc-cifar10}).

\begin{figure}
  \centering
    \includegraphics[width=0.9\textwidth]{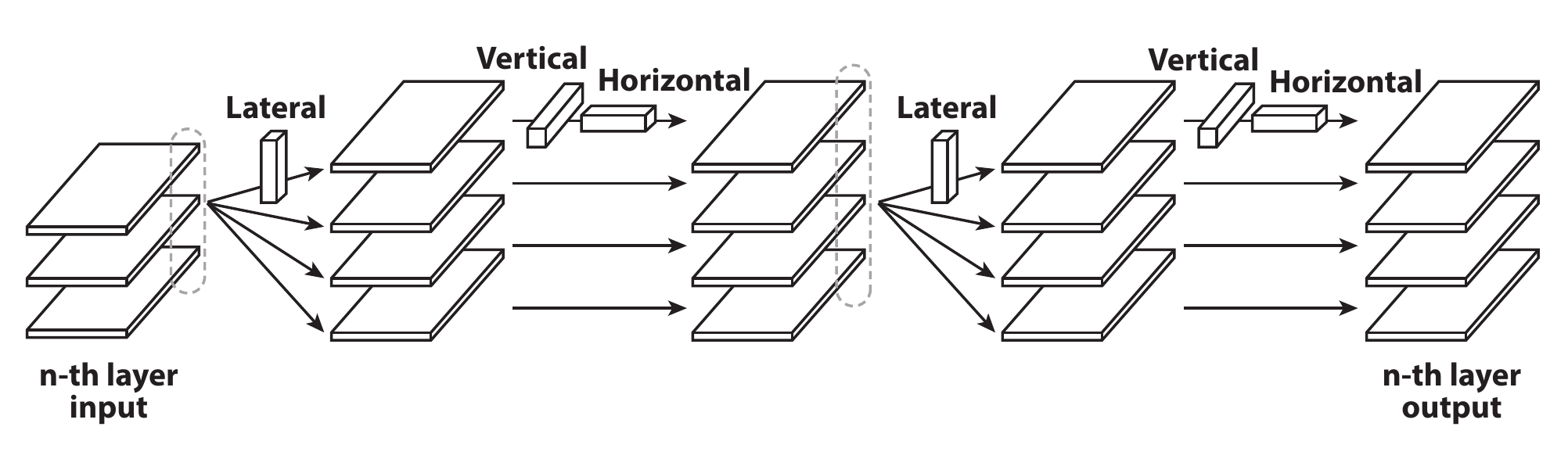}
    \caption{A single layer structure of flattened convolutional networks.
             Flattened layer includes $l$ sets of 1D-separated convolutions over channels (lateral, $L$), vertical ($V$) and horizontal ($H$) direction.
             In this work, two stages of $LVH$ combinations ($l=2$) were chosen by cross-validation and it reported the same accuracy as measured in baseline model.
	     $V$ and $H$ convolutions are operated in \textit{full} mode to preserve the same output dimension as the baseline model.
             Bias is added after each of 1D convolution, but skipped in this illustration.
             No non-linear operator is applied within the flattened layer.}
     \label{fig:illust-network}
\end{figure}

\subsection{Training Flattened Model}
In flattened model, we use CNNs constructed with 1D filters as described in Figure \ref{fig:comparison-filters}.
First, lateral filters ($L$) perform convolution across channels like \textit{mlpconv} layers in Network in Network \citep{lin_2013_nin}.
Then, each channel is convolved with vertical and horizontal filters ($V$ and $H$) whose filter sizes are $Y \times 1$ and $1 \times X$ in space respectively.
While \textit{mlpconv} applies 1D convolution across channels, this work extends 1D convolutions in space as well.
Thus, the proposed method can be viewed as a generalization of training with 1D-separated filters in $\mathbb{R}^3$.
Once the model structure is defined at the training stage, no post processing or fine-tuning is needed.
The structural constraint forces the model to learn 1D separated filters, equivalently a rank-one 3D filter, except biases.

Replacing filters with dimension of ${C \times X\times Y}$ to filters dimensions of $C$, $X$, $Y$ resulted in $2-3\%$ accuracy drops in our experiments on different datasets.
1D separated filters with dimensions of $C$, $X$ and $Y$ contains only $5\%$ of the parameters as in 3D filter in commonly used CNNs \citep{krizhevsky2012imagenet, sermanet_overfeat2013}.
While \citet{denil2013predicting} demonstrates that $5\%$ of essential parameters can predict the rest of parameters in the best case, such reduction in parameters accompanies accuracy loss \citep{gong_compress_vq2014, lebedev_speedup_cp2014}.
In our 1D training pipeline, parameters in one set of 1D filters are not enough to distinguish discriminate features, thus, results in failure to recover the target performance.
We found that removing one direction from the $LVH$ pipeline caused a significant accuracy loss, which implies that convolutions toward all directions are essential.

We cascaded a set of Lateral-Vertical-Horizontal ($LVH$) convolutional layers to compensate accuracy loss at the cost of using more parameters.
Empirical results show that two cascaded sets of $LVH$ layers achieve the same performance as a standard convolution layer in the baseline model.
The flattened layer, consisting of two stages of $LVH$, is illustrated in Figure \ref{fig:illust-network} and used throughout the experiments.
The $V$ and $H$ convolutions within the flattened layer are operated in \textit{full} mode in order not to lose boundary features and to keep output dimensions the same as the baseline model.
With this method, we were able to decrease the number of parameters $8-10\times$ compared to that in the baseline model.
Different number of $LVH$ layers could be cascaded depending on the difficulties of classification tasks and the effect of parameter reduction is discussed in the section \ref{sec:param-reduction}.
We used the same training configuration of baseline model except decreased weight decay to give more freedom to adapt features.

The serialized model with 1D convolutions is more vulnerable to vanishing gradient problem than standard CNNs.
Attaching many 1D convolution layer provides more flexibility in filter shape therefore helps draw delicate decision boundary.
However, longer gradient path experiences more steps of parameter updates and error accumulation, which possibly cause fast decaying gradients.
This trend is more persistent for $V$ or $H$ convolution because they get feedback from only one channel.
From the gradients update $\delta x_{i} = \sum_{k=1}^K w_{k} \delta x_{i+1}$, accumulation of gradients and attenuation by weights are balanced each other in the standard CNNs.
However, the flattened structure has generally few connections in $H$ and $V$ convolution.
Vanishing gradients can be handled with smart weight initialization.
Normalized initialization balanced with forward and backward passes \citep{glorot2010weight} or constant error propagation helps deliver gradients to lower layers.
In our experiments, they yielded more successful training than the heuristic \citep{lecun_1998_backprop} while the baseline model with the highest accuracy was trained with the heuristic initialization.

\begin{figure}
  \centering
  \begin{subfigure}[b]{0.5\textwidth}
    \includegraphics[width=\textwidth]{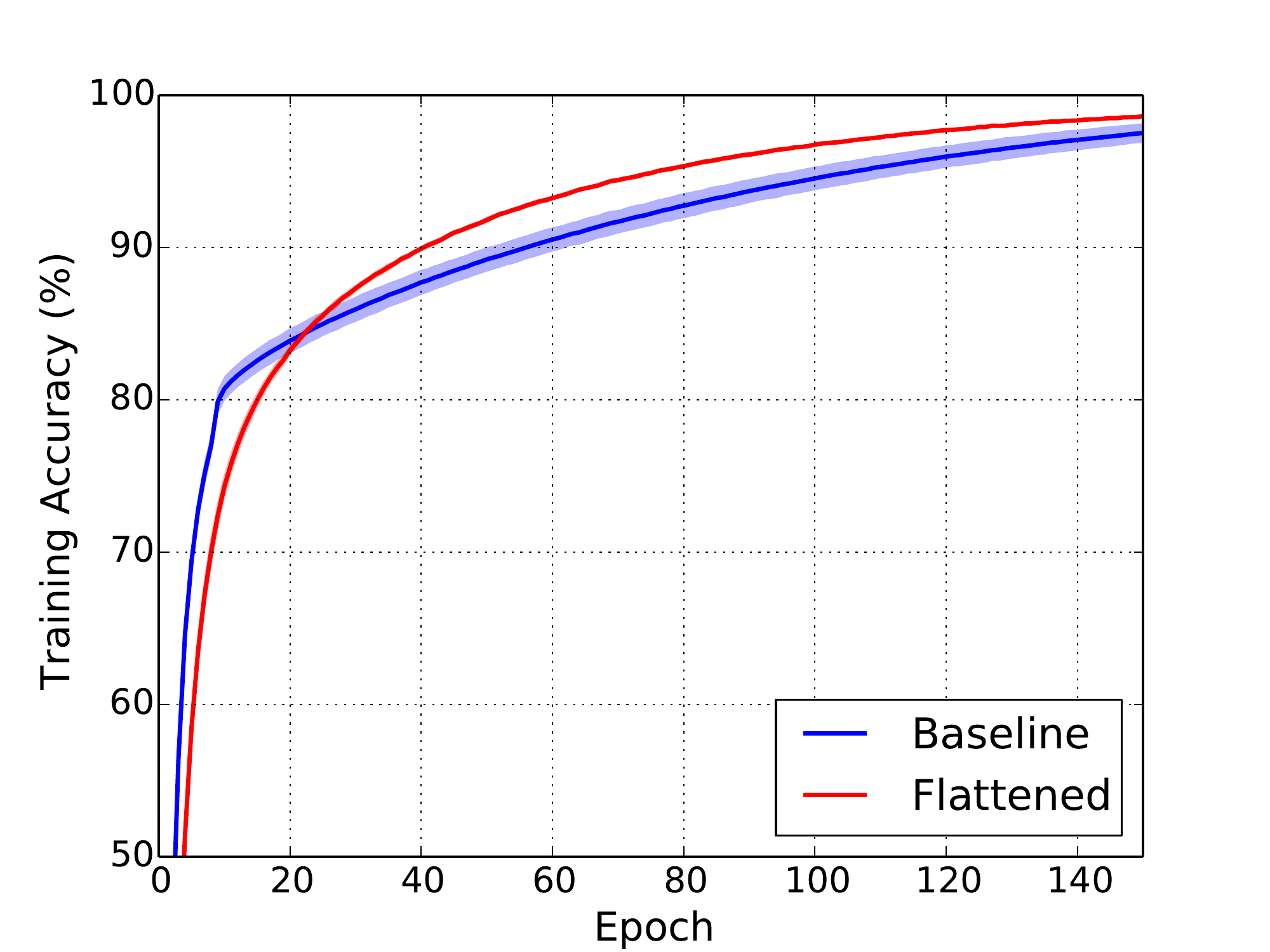}
    \caption{Training accuracy}
    \label{fig:convergence-base}
  \end{subfigure}%
  \begin{subfigure}[b]{0.5\textwidth}
    \includegraphics[width=\textwidth]{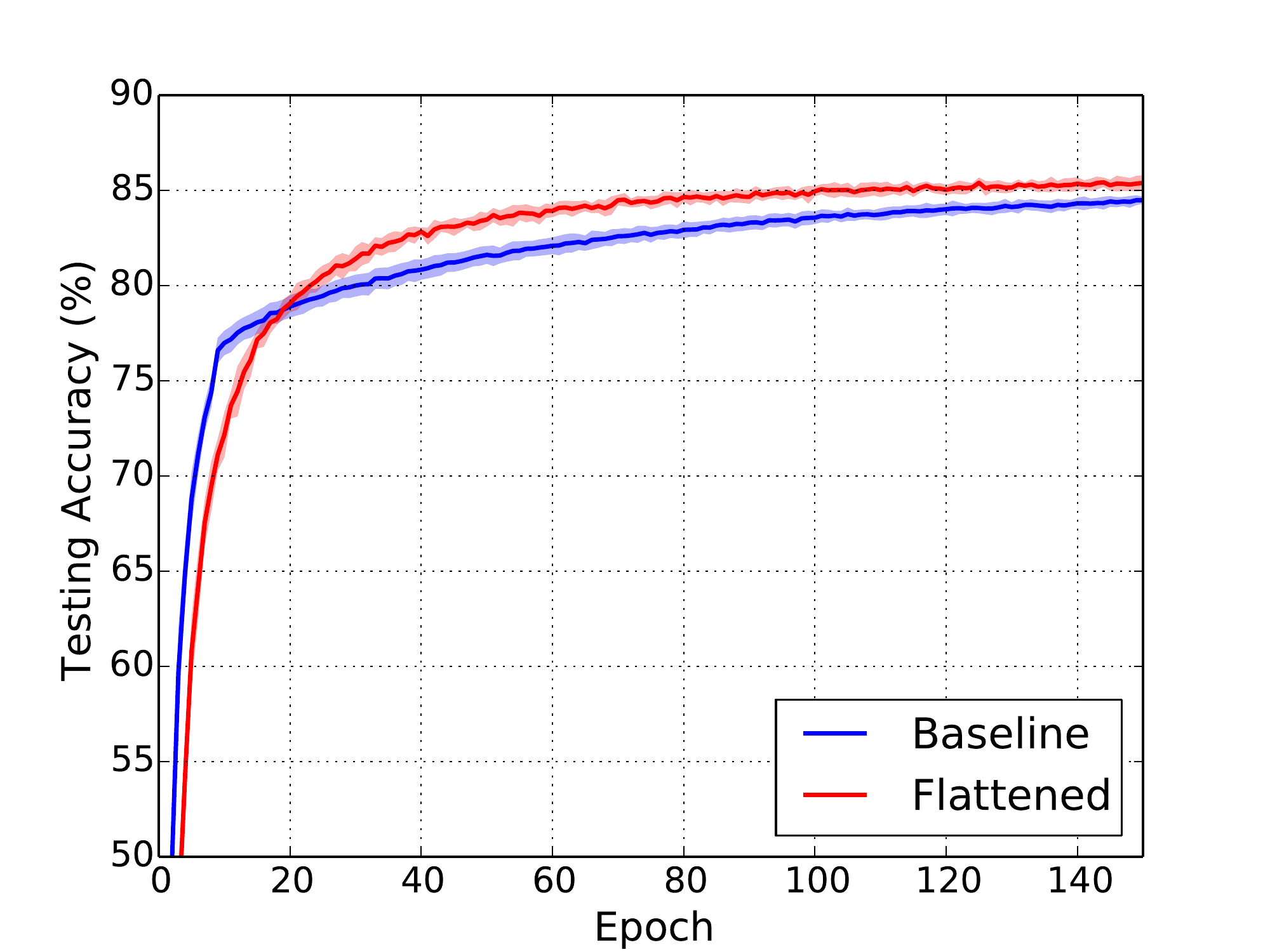}
    \caption{Testing accuracy}
    \label{fig:convergence-flat}
  \end{subfigure}
  \caption{
    Convergence rate of flattened and baseline models both in training and testing.
    The structure of baseline and flattened model is specified in the section \ref{sec:train-baseline} and the table \ref{table:model-spec-cifar10}, respectively, and CIFAR-10 dataset is used in this experiment.
    The solid line denotes a mean and the shade around the line indicates a standard deviation of the curve.
    The variation of training for the flattened model is too small to appear in the illustration.
  }
  \label{fig:convergence}
\end{figure}

With the proper weight initialization, the flattened model achieves the comparable accuracy as the baseline model.
Figure \ref{fig:convergence} reports the training and testing accuracy on the CIFAR-10 dataset.
Considering that learning rate for the baseline model is decreased in order to achieve the highest accuracy, the baseline model saturates earlier and to a lower accuracy compared to the flattened model.
The shaded region in the plots indicates a variation of accuracies obtained from samples with different random seeds.
The learning curve of the flattened model shows consistent results with less variation.
The flattened method alleviates the training effort by accelerating backpropagation and is presented in the section \ref{sec:profiling}.

\begin{figure}
  \centering
  \begin{subfigure}[b]{0.45\textwidth}
    \includegraphics[width=\textwidth]{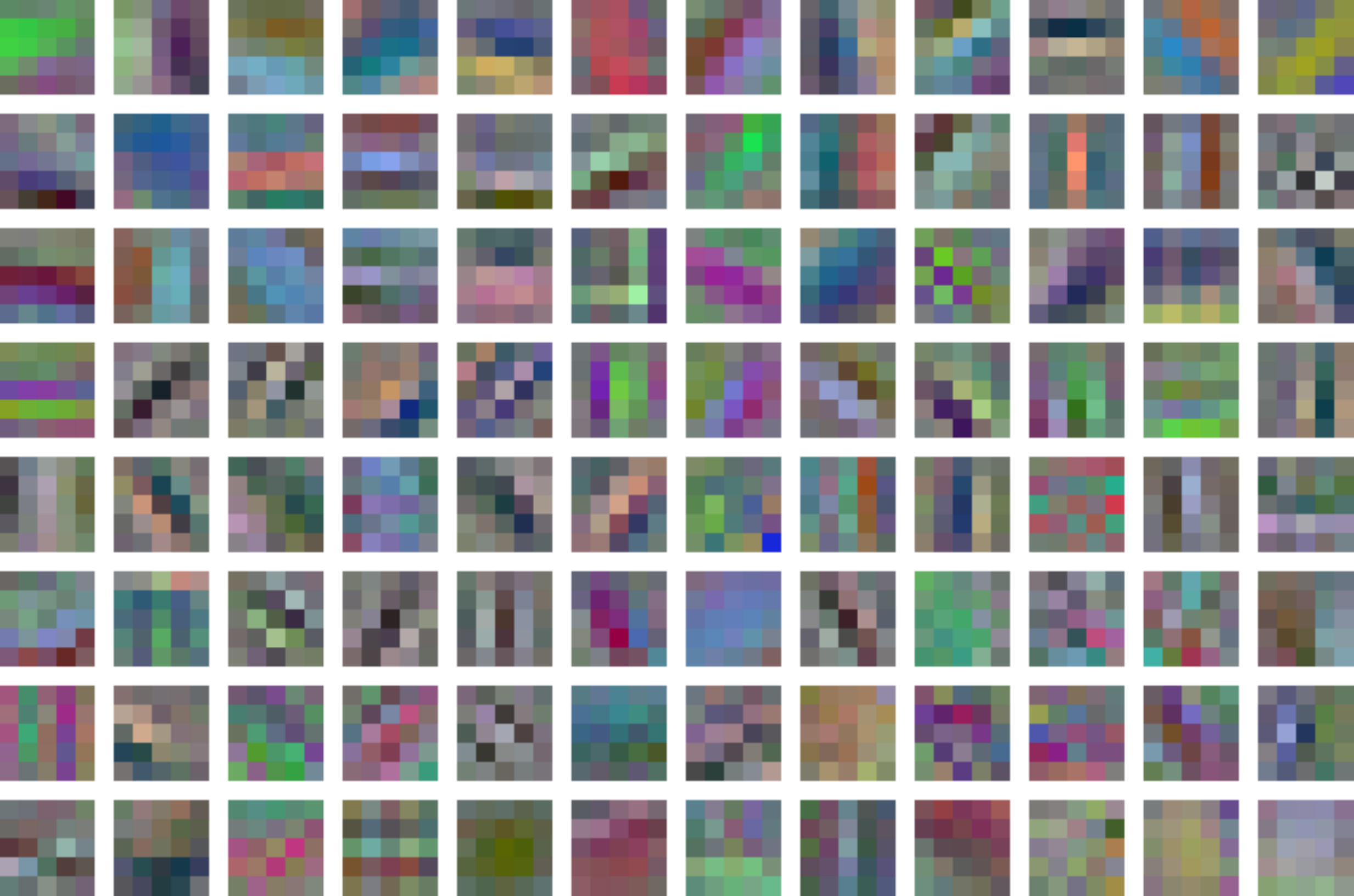}
    \caption{Filters in baseline model}
    \label{fig:filter-reference}
  \end{subfigure}%
  \qquad
  \begin{subfigure}[b]{0.45\textwidth}
    \includegraphics[width=\textwidth]{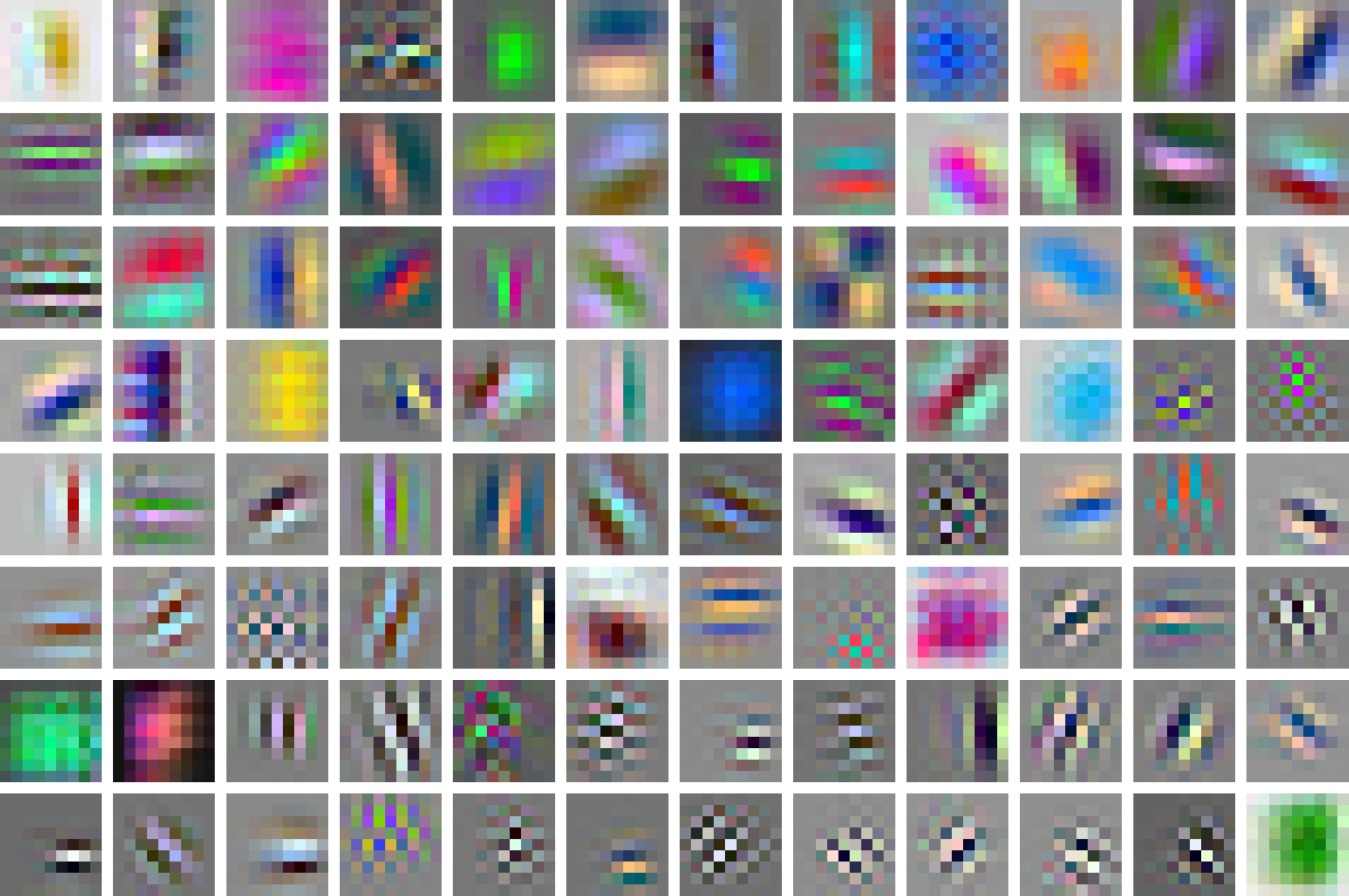}
    \caption{Reconstructed filters in flattened model}
    \label{fig:filter-flatten}
  \end{subfigure}
  \caption{Visualization of the first layer filters trained on CIFAR-10.
           Filters are reconstructed from cross-product of 1D convolution filters and contains clear and diverse features.
           Filters are sorted by variance in descending order and bias is excluded during reconstruction.}
  \label{fig:filters}
\end{figure}

The first layer filters trained on CIFAR-10 within the flattened structure are reconstructed to 3D filters and presented in Figure \ref{fig:filters} though the first layer is not converted in the main experiment (see Section \ref{sec:param-reduction}).
Filters are reconstructed by cross-product of 1D filters trained from two sets of $LVH$ convolution layers.
The richness of distinct features in the filters is necessary for model discriminability.
Surprisingly, the reconstructed filters have distinct Gabor edges and color blobs.
Features have high contrast and edges are sparse as in \citet{lee_2007_sparse_code} without $L_1$ penalty.
This finding supports the effectiveness of the method and explains the comparable performance of proposed model over the baseline model.

\subsection{Parameter Reduction}
\label{sec:param-reduction}

Flattened model applied to CNNs generally relaxes computational demands by reducing the number of parameters.
Here we analyze parameter reduction and its trade-off in practical viewpoint.

Flattened convolutional layer used in this work has two stages of $LVH$ convolutions.
Corresponding filter dimensions is $F(F + C + 2X + 2Y)$ while the regular 3D filters in the baseline model has dimension of ${CXYF}$, where $C$ and $F$ denote the number of input and output channels, $X$ and $Y$ are the spatial dimensions of the filter.

Considering the filter dimensions are relatively small and fixed throughout layers, our gain for parameter reduction is mostly depend on the ratio between $F$ and $C$.
We denote the ratio $k = F/C$ and we want to flatten a convolutional layer only if $k$ satisfies the equation \ref{eq:reduction-interval}
\begin{equation}
  C^2k^2 + C^2k + 2C(X + Y) < C^2XYk
  \label{eq:reduction-interval}
\end{equation}
where the left side denotes the computions required for a flattened layer and the right side for a standard convolution layer.

\begin{wrapfigure}{r}{0.5\textwidth}
  \vspace{-25pt}
  \begin{center}
    \includegraphics[width=0.5\textwidth, trim = 0mm 0mm 15mm 0mm]{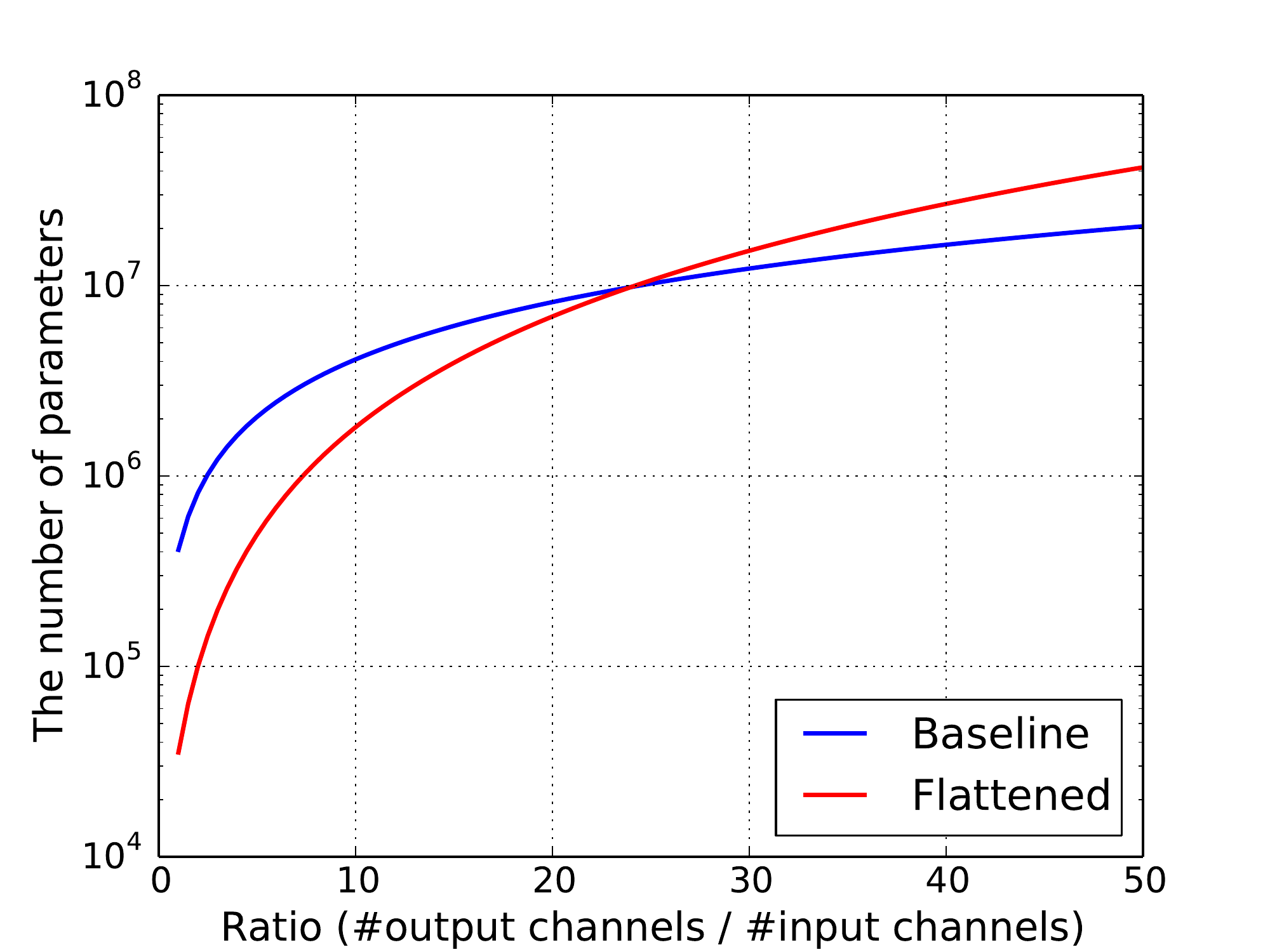}
  \end{center}
  \vspace{-10pt}
  \caption{
           Filter dimensions of $X=Y=5$ and $C = 128$ are used to observe reduction efficiency.}
  \label{fig:tradeoff}
  \vspace{-10pt}
\end{wrapfigure}

Figure \ref{fig:tradeoff} visualizes the relationship between the parameter reduction of flattened network compare to baseline network.
The flattening method is guaranteed to reduce a large portion of parameters as long as the number of channels increases smoothly.
Most layers of CNNs can benefit from this method since the number of channels does not decrease over layers and the ratio between channels usually resides between 1 and 3, other than the first layer \citep{krizhevsky2012imagenet, sermanet_overfeat2013, szegedy_2014_inception, simonyan2014very}.
First layer begins with three channels, so performing $96$ filters gives us a ratio of $32$ where the baseline model has less parameters.
For the purpose of parameter reduction, we selectively applied flattening to the second and third layers of the baseline CNN though we successfully applied the flattening to the all convolutional layers and achieved the same accuracy.
Table \ref{table:model-spec-cifar10} summarizes details of two CNN models in terms of the number of parameters.


\begin{table}[h]
  \caption{The number of parameters in each layer of flattened and baseline CNN model.}
  \label{table:model-spec-cifar10}
  \begin{center}
    \begin{tabular}{l r r r r}
      \multicolumn{1}{c}{\bf } & \multicolumn{1}{c}{\bf Baseline Model} &
      \multicolumn{1}{c}{\bf Flattened Model} &\multicolumn{1}{c}{\bf Reduction}
      \\ \hline \\
      Layer $1$ Parameters &   $7,200$ & Not applied
        \tablefootnote{The flattening is not applied to the first layer for purpose of parameter reduction based on figure \ref{fig:tradeoff} though the technique in the first layer achieved comparable accuracy as well.}
                                                     &  $0.0\%$ \\
      Layer $2$ Parameters & $307,200$ &    $30,912$ & $89.9\%$ \\
      Layer $3$ Parameters & $819,200$ &   $102,144$ & $87.5\%$ \\
    \end{tabular}
  \end{center}
\end{table}

\subsection{Memory Usage}

\begin{wrapfigure}{r}{0.5\textwidth}
  \vspace{-25pt}
  \begin{center}
    \includegraphics[width=0.5\textwidth, trim = 0mm 0mm 15mm 0mm]{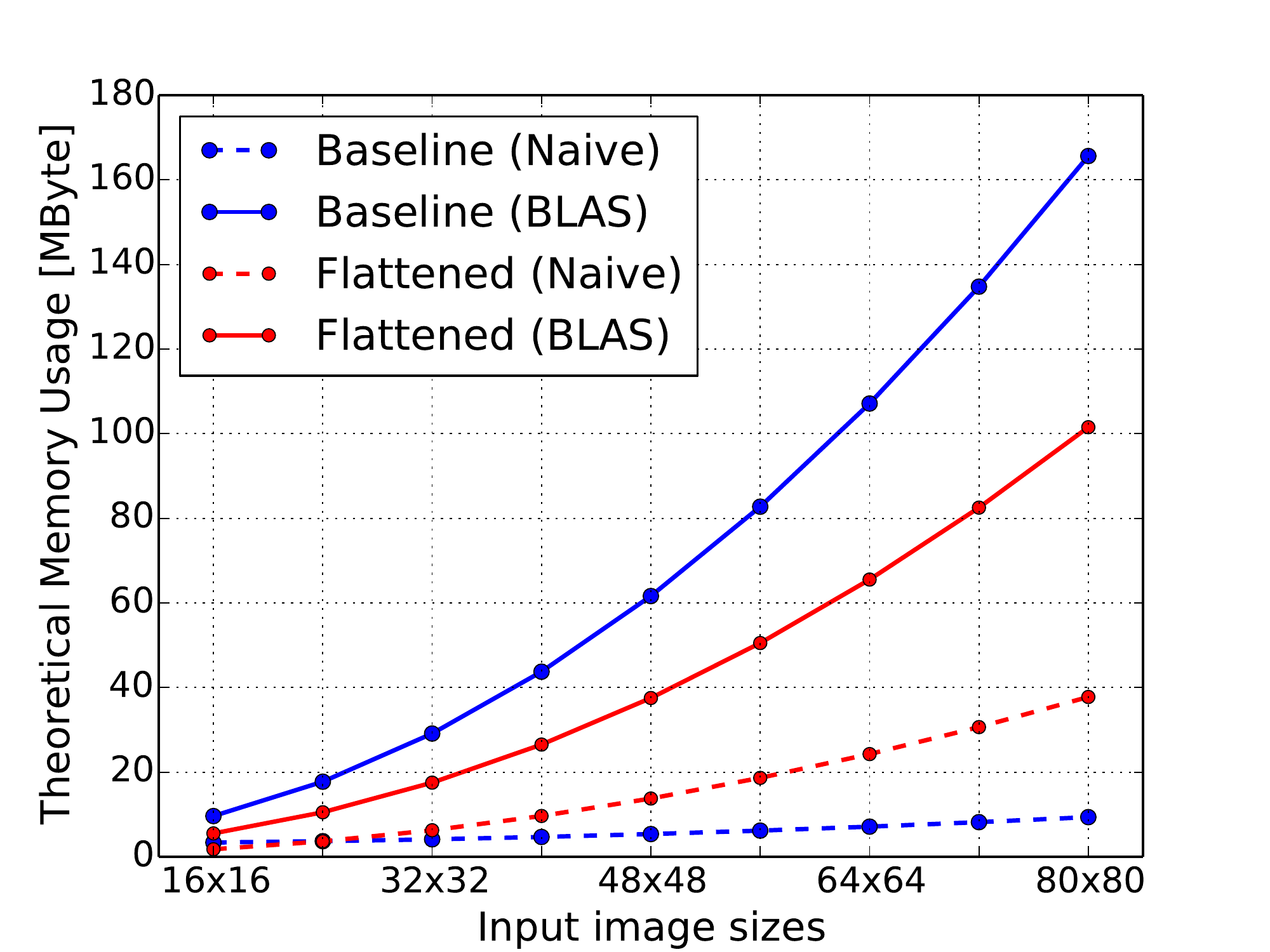}
  \end{center}
  \vspace{-10pt}
  \caption{
    Theoretical memory usage for a single flattened layer.
    Memory usage optimized implementation (Na{\"i}ve) and performance optimized implementation (BLAS) are presented.
  }
  \label{fig:memory-usage}
  \vspace{-14pt}
\end{wrapfigure}

In this section we compare the memory consumption of the baseline and flattened network in training.
It is an important concern since memory limit could affect the degree of parallelism as the model scales up.
The flattened layer needs to hold all intermediate states for the backward pass whose size is as big as the size of output planes.
Considering that each convolution is broken down into $N$ pieces of 1D convolution, the flattened structure produces $N-1$ intermediate states that needed to be stored in the memory.
This is true for the na{\"i}ve implementation of convolution as seen in figure \ref{fig:memory-usage} where it uses nested for-loops and is optimized in memory usage.

However, BLAS-friendly convolution routine is generally used to achieve the highest performance in time and adapted in scientific computing framework \citep{collobert_torch7_2011, jia2014caffe, chetlur2014cudnn}.
The na{\"i}ve approach exploits limited parallelism due to its frequent memory access, which throttles the real-time performance of CNNs on many platforms.
The BLAS implementation for the baseline model handles convolution as a matrix multiplication at the cost of additional memory copy and space.
On the other hand, 1D convolution pipeline in flattened layer can achieve sufficient parallelism without additional resources in feedforward and with little extra memory in backpropagation as opposed to the 3D convolution.
Therefore, the flattened layer uses less memory than a baseline convolution layer in practice even though each convolution is broken down into 6 pieces as illustrated in figure  \ref{fig:memory-usage}.

\subsection{Classification Accuracy}
Despite of the reduced number of parameters, the flattened model does not suffer from accuracy loss.
We tested the model on CIFAR-10, CIFAR-100 and MNIST datasets.

The CIFAR-10 and CIFAR-100 \citep{krizhevsky2009learning} are datasets of $32\times32$ RGB images and often used as a standard measurement to evaluate classification capability.
CIFAR-10 has 10 classes while CIFAR-100 has 100 classes.
Both datasets consist of with $50,000$ training and $10,000$ testing images.
Before use, datasets are preprocessed with contrast normalization per image followed by ZCA whitening to the whole dataset as \citet{goodfellow2013maxout}.
Both models tend to reach the same performance but in our experiments the accuracy of flattened model outperforms the baseline model slightly as can be seen from Table \ref{table:acc-cifar10}.

The MNIST dataset \citep{lecun1998gradient} consists of hand written digits of 0-9.
This dataset contains $60,000$ training and $10,000$ testing images.
We applied contrast normalization per image as \citet{goodfellow2013maxout} without ZCA whitening since hand-written images have low cross-correlation values as opposed to CIFAR datasets.
The classification on MNIST is relatively easier than CIFAR datasets and the accuracies of two models are highly saturated.
Whereas it is difficult to compare models because of the simplicity of the dataset, baseline and flattened models give almost the same accuracies.

\begin{table}[h]
  \caption{Classification accuracy of baseline and flattened model on different datasets}
  \label{table:acc-cifar10}
  \begin{center}
    \begin{tabular}{l l c c}
      \multicolumn{1}{c}{\bf Dataset}  &\multicolumn{1}{c}{\bf Model Type}  &\multicolumn{1}{c}{\bf Test Accuracy} \\
       \hline \\
       \multirow{2}{*}{CIFAR-10}   &Baseline Model  & $86.42\%$ \vspace{2pt}\\
                                   &Flattened Model & $87.04\%$ \vspace{5pt}\\
       \hline \\
       \multirow{2}{*}{CIFAR-100}  &Baseline Model  & $60.08\%$ \vspace{2pt}\\
                                   &Flattened Model & $60.92\%$ \vspace{5pt}\\
       \hline \\
       \multirow{2}{*}{MNIST}      &Baseline Model  & $99.62\%$ \vspace{2pt}\\
                                   &Flattened Model & $99.56\%$ \vspace{5pt}\\
      \hline
    \end{tabular}
  \end{center}
\end{table}

\subsection{Acceleration}
\label{sec:profiling}

\begin{figure}[!ht]
  \centering
  \begin{subfigure}[b]{0.5\textwidth}
    \includegraphics[width=\textwidth]{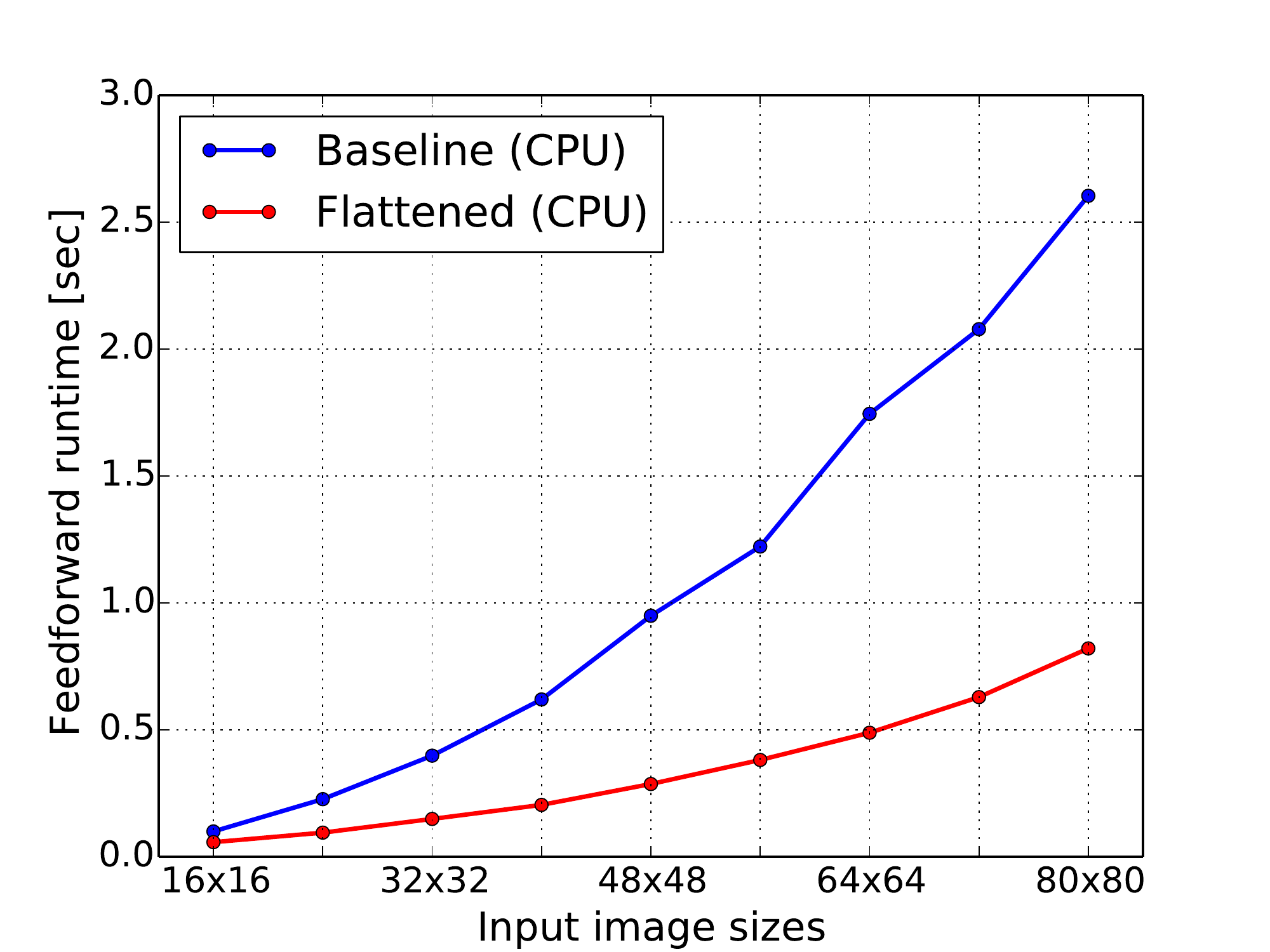}
    \caption{Feedforward on CPU}
    \label{fig:forward-CPU}
  \end{subfigure}%
  \begin{subfigure}[b]{0.5\textwidth}
    \includegraphics[width=\textwidth]{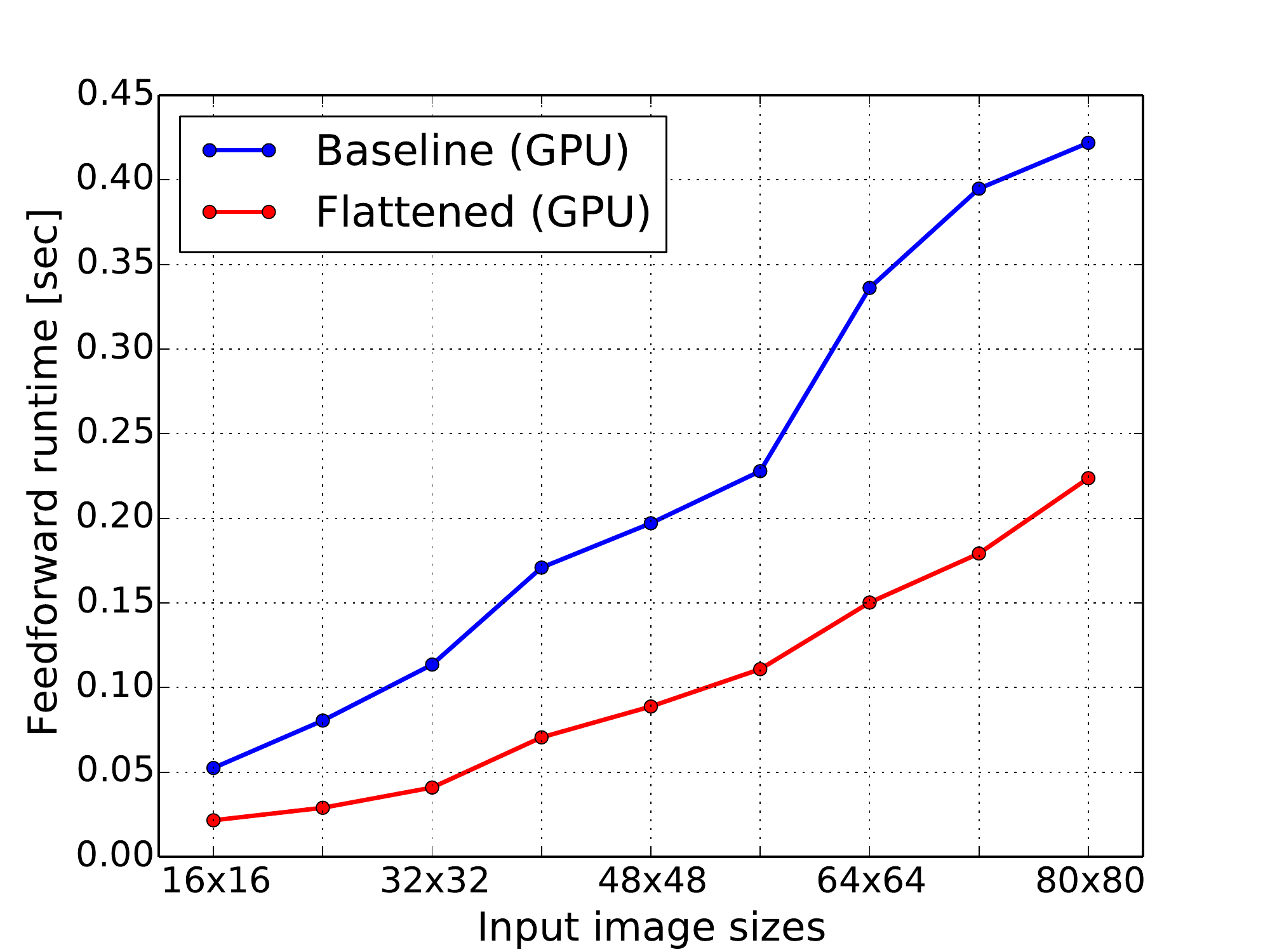}
    \caption{Feedforward on GPU}
    \label{fig:forward-GPU}
  \end{subfigure}
  \begin{subfigure}[b]{0.5\textwidth}
    \includegraphics[width=\textwidth]{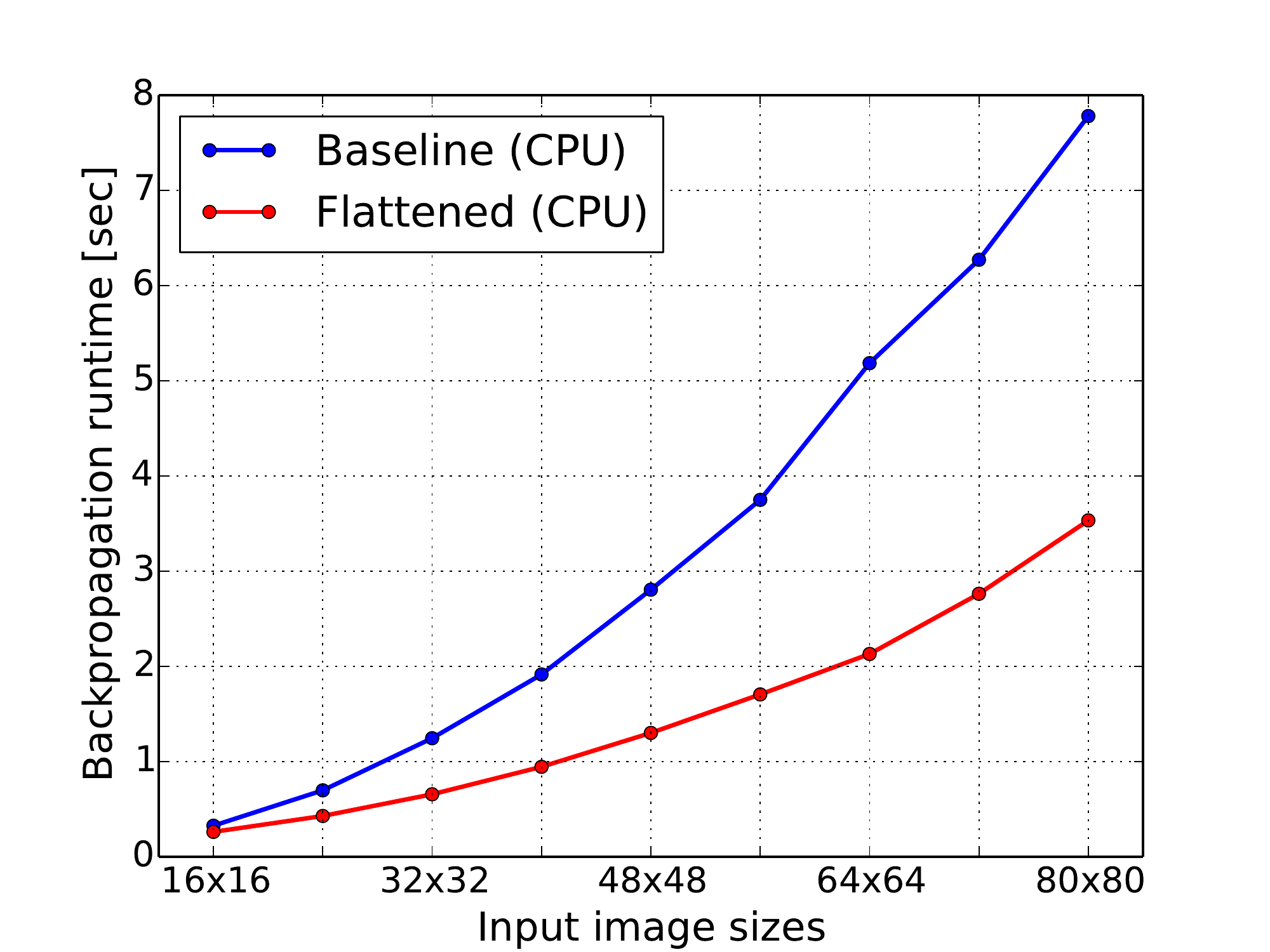}
    \caption{Backpropagation on CPU}
    \label{fig:backward-CPU}
  \end{subfigure}%
  \begin{subfigure}[b]{0.5\textwidth}
    \includegraphics[width=\textwidth]{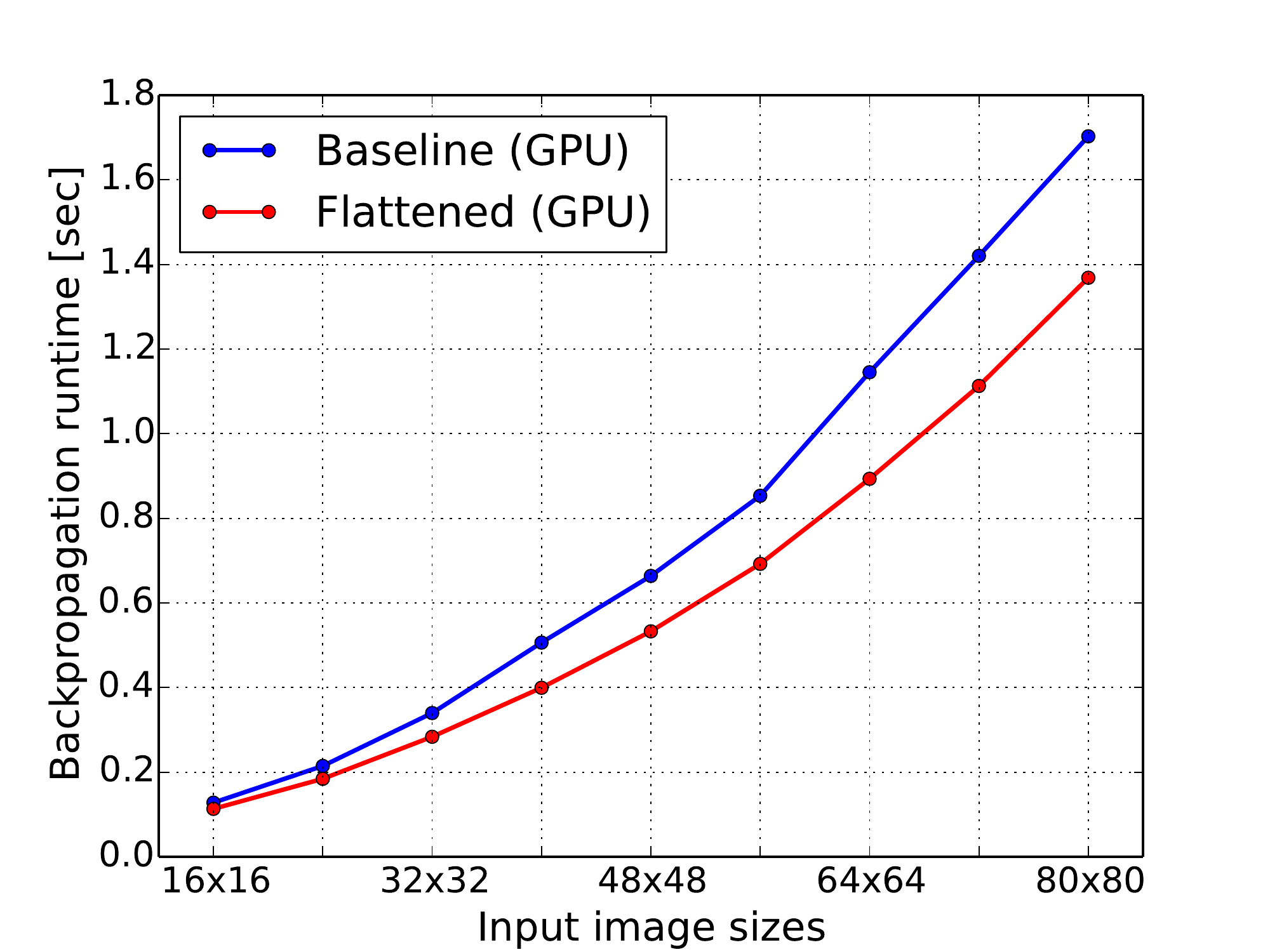}
    \caption{Backpropagation on GPU}
    \label{fig:backward-GPU}
  \end{subfigure}
  \caption{Speed-up comparison for feedforward and backpropagation execution time measured on CPU and GPU.
           Input images with $128$ batch size were applied to a single flattened layer,
           which is equivalent to $128$-to-$128$ convolution with $5\times5$ filter in terms of classification accuracy.
           Flattened model runs faster on both platforms and the performance gap increases as the size of image is larger.
           Profiling results were measured on BLAS-accelerated convolution routines for all experiments.
           More optimized CUDA implementation for 1D convolution would shorten execution time on GPU.}
  \label{fig:evaluation}
\end{figure}

Due to the reduced parameters in 1D convolution pipelines, flattened structure reduces computational demands of CNNs thus accelerate both feedforward and backward computations of CNNs.
Profiling results of flattened and baseline models for feedforward and backpropagation passes are presented in Figure \ref{fig:evaluation}.
We ran the same tests on CPU and GPU to check performance of flattened structure versus different degree of parallelism.
The performance was measured on Intel i7 3.3GHz CPU and NVIDIA Tesla K-40 GPU.
Different image sizes from $16\times16$ to $80\times80$ with $128$ batch were applied to the models.
We used a single convolution layer for baseline model which has dimensions of $128\times128\times5\times5$ filters, and the corresponding 6 pieces of convolution layer for flattened model which has 2 sets of filters with dimensions of $128\times128\times1\times1$, $128\times128\times5\times1$ and $128\times128\times1\times5$ filters.

Feedforward pass is the part that mostly benefits from the manipulation of 1D convolution filters.
The flattened layer runs about two times faster than the conventional convolutional layer during forward pass on both platforms.
The acceleration tends to increase as the size of images gets larger because overhead time becomes negligible for large size of images.
We implemented $L$, $V$ and $H$ convolution routines that are optimized in speed.
It is evident that the filter separation reduces the absolute number of computations, which provides the acceleration.
Efficient memory access is another factor for feedforward acceleration.
In the 1D convolution pipeline, convolution is processed along one direction only, therefore input pixels can be considered as a 1D image sequence, which minimizes effort in data index and access.
The results imply that the flattened structure is more promising if used in the lower layers of CNNs where image size is not small and the ratio between input/output channels increases smoothly.

Using flattened layer also reduces training time on CPU and GPU.
The backpropagation consists of gradient propagation and parameter update.
In the former case, its computation benefits from the coalesced memory access of 1D convolution pipeline as in feedforward.
However, the latter requires data accumulation over all pixels at each 1D convolutional layer.
Therefore the serial operation with frequent global memory access make the acceleration negligible on GPU.

\section{Conclusion}

In this work, we propose a flattening technique to reduce the number of parameters in convolutional neural networks for feedforward acceleration.
We convert  each of convolutional layer of 3D convolution pipeline into a sequence of 1D convolutions across channels, vertical and horizontal directions in training phase.
We found that the model molded in 1D structure is able to learn 1D filters successfully and achieves about two times speed-up in evaluation compared to the baseline model.
Furthermore, with ten times less parameters, the flattened convolutional networks achieve similar or better accuracies on CIFAR-10, CIFAR-100, and MNIST.
In addition, the proposed method does not require efforts in manual tuning or post processing once the model is trained and it bypasses the difficulties in solving optimization problem to learn 1D filters.
The simple nature of the parameter reduction method could be applied to accelerate a very large-scale model as well and this remains as future work.

\subsubsection*{Acknowledgments}

This work is supported by Office of Naval Research (ONR) grants 14PR02106-01 P00004 and MURI N000141010278.
We gratefully appreciate the support of NVIDIA Corporation with the donation of GPUs used for this research.

\bibliography{iclr2015}
\bibliographystyle{iclr2015}

\end{document}